\documentclass[conference]{IEEEtran}

\usepackage{amsmath,amsfonts,bm}

\def\eqref#1{equation~\ref{#1}}
\def\1{\bm{1}}

\def\vb{{\bm{b}}}

\def\vh{{\bm{h}}}

\def\vr{{\bm{r}}}

\def\vx{{\bm{x}}}
\def\vy{{\bm{y}}}
\def\vz{{\bm{z}}}

\def\mH{{\bm{H}}}
\def\mI{{\bm{I}}}

\def\mR{{\bm{R}}}

\def\mU{{\bm{U}}}
\def\mV{{\bm{V}}}

\def\mX{{\bm{X}}}
\def\mY{{\bm{Y}}}

\DeclareMathAlphabet{\mathsfit}{\encodingdefault}{\sfdefault}{m}{sl}
\SetMathAlphabet{\mathsfit}{bold}{\encodingdefault}{\sfdefault}{bx}{n}
\newcommand{\tens}[1]{\bm{\mathsfit{#1}}}

\def\tW{{\tens{W}}}
\def\tX{{\tens{X}}}

\newcommand{\KL}{D_{\mathrm{KL}}}

\IEEEoverridecommandlockouts
\usepackage{cite}
\usepackage{amsmath,amssymb,amsfonts}
\usepackage{algorithmic}
\usepackage{graphicx}
\usepackage{textcomp}
\usepackage{xcolor}

\usepackage{hyperref}
\usepackage{url}
\usepackage{multirow}
\usepackage{here}
\usepackage{bm}
\usepackage{algorithm}
\usepackage{algorithmic}

\newcommand\blfootnote[1]{%
  \begingroup
  \renewcommand\thefootnote{}\footnote{#1}%
  \addtocounter{footnote}{-1}%
  \endgroup
}

\def\BibTeX{{\rm B\kern-.05em{\sc i\kern-.025em b}\kern-.08em
    T\kern-.1667em\lower.7ex\hbox{E}\kern-.125emX}}
\begin{document}

\title{Multi-Decoder RNN Autoencoder Based on Variational Bayes Method
%\\
}

\author{\IEEEauthorblockN{ Daisuke Kaji}
\IEEEauthorblockA{\textit{AI R \& D Division} \\
\textit{Denso Corporation}\\
Tokyo, Japan \\
daisuke.kaji.j3a@jp.denso.com}
\and
\IEEEauthorblockN{ Kazuho Watanabe}
\IEEEauthorblockA{\textit{Dept. of CSE} \\
\textit{Toyohashi University of Technology}\\
Aichi, Japan \\
wkazuho@cs.tut.ac.jp}
\and
\IEEEauthorblockN{Masahiro Kobayashi}
\IEEEauthorblockA{\textit{Dept. of CSE} \\
\textit{Toyohashi University of Technology}\\
Aichi, Japan \\
kobayashi@lisl.cs.tut.ac.jp}
}

%\author{\IEEEauthorblockN{1\textsuperscript{st} Daisuke Kaji}
%\IEEEauthorblockA{\textit{AI R &D Division} \\
%\textit{Denso corporation}\\
%Tokyo, Japan \\
%daisuke.kaji.j3a@jp.denso.com}
%\and
%\IEEEauthorblockN{2\textsuperscript{nd} Kazuho Watanabe}
%\IEEEauthorblockA{\textit{Department of Computational Neuroscience} \\
%\textit{Toyohashi University of Technology}\\
%Aichi, Japan \\
%wkazuho@cs.tut.ac.jp}
%\and
%\IEEEauthorblockN{3\textsuperscript{rd} Masahiro Kobayashi}
%\IEEEauthorblockA{\textit{Department of Computational Neuroscience} \\
%\textit{Toyohashi University of Technology}\\
%Aichi, Japan \\
%kobayashi@lisl.cs.tut.ac.jp}
%}

\maketitle

\begin{abstract}
Clustering  algorithms have wide applications and play an important role in 
data analysis fields including time series data analysis.
However, in time series analysis, most of the algorithms used signal 
shape features or the initial value of hidden variable of a neural network.
Little has been discussed on the methods based on the generative model of the time 
series.
In this paper, we propose a new clustering algorithm focusing on the generative process
of the signal with a recurrent neural network and the variational Bayes method.
Our experiments show that the proposed algorithm not only has a robustness against for phase shift, amplitude and signal length variations but also provide a flexible clustering  based on the property of the variational Bayes method.
\end{abstract}


%for arXiv
\blfootnote{$\copyright$ 2020 IEEE. Personal use of this material is permitted. Permission from IEEE must be obtained for all other uses, in any current or future media, including
reprinting/republishing this material for advertising or promotional purposes, creating new collective works, for resale or redistribution to servers or lists, or
reuse of any copyrighted component of this work in other works. DOI: XX.XXXX/IJCNN.2020.XXXXXX.}

%\section{Introduction}
\section{Introduction}
The rapid progress of IoT technology has brought huge data in wide fields such as traffic, industries, medical research  and so on.
Most of these data are gathered continuously and accumulated as time series data, and the extraction of features from a time series have been studied intensively in recent years.
The difficulty of time series analysis is the variation of the signal in time which  gives rise to phase shift, compress/stretch and length variation. 
Many methods have been proposed to solve these problems.
Dynamic Time Warping (DTW) was designed to measure the distance between warping signals \cite{DTW}.
This method solved the compress/stretch problem by applying a dynamic planning method.
Fourier transfer or wavelet transfer can extract the features based on the frequency components of signals.
The phase shift independent features are obtained by calculating the power spectrum of the transform result.

In recent years, the recurrent neural network (RNN), which has recursive neural network structure, has been  widely used in time series 
analysis \cite{RNN_Elman1,RNN_Elman2}. This recursive network structure makes it possible to retain the past information of time series. Furthermore, this architecture enables us to apply this algorithm to signals with different lengths.
Although the methods mentioned above are effective solutions for the compress/stretch, phase shift and signal length variation issues, little has been studied about these problems comprehensively.

Let us turn our attention to feature extraction again. Unsupervised learning using a neural network architecture autoencoder (AE) has been studied as a feature extraction method \cite {AE_Hinton1,Pascal, Salah}. 
AE using RNN structure (RNN-AE)  has also been proposed \cite{Srivastava} and it has been applied to real data such as driving data \cite{Dong} and others.  
RNN-AE can be also interpreted as the discrete dynamical system: chaotic behavior and the deterrent method have been studied from this point of view \cite{Zerroug,Laurent}.

In this paper,  we propose a new clustering algorithm for feature extraction focusing on the dynamical system aspect of RNN-AE.
In order to achieve this, we employed a multi-decoder AE to describe different dynamical systems as a generative model. 
We also applied the variational Bayes method \cite{Attias,Ghahramani,kaji} as the clustering algorithm.

This paper is composed as follows:
in Section \ref{AE}, we explain AE  from a dynamical system view, then we define our model and from this, derive its learning algorithm. In Section \ref{Exp}, we describe the application of  our algorithm to an actual time series to show its robustness, including experiments 
using periodic data, complex periodic data and driving data. 
Finally we summarize our study and describe our future work in Section \ref{Summary}.

\section{Related work}
A lot of excellent clustering/representation algorithms of data using AE have been studied so far \cite{Tschannen}. 
Song et al. \cite{Song} integrated the 
distance between data and centroids into an objective function to obtain a cluster structure in the encoded data space. 
Pineau and Lelarge \cite{Pineau}  proposed a generative model based on the variational autoencoder (VAE) \cite{Diederik} with a clustering structure as a prior distribution, VAE was also applied to the hierarchical clustering method of time series data \cite{Per}.
Wang et al. \cite{Wang} achieved a high separability clustering result by adding a regularization term for 
the orthogonality and balanced clusters of the encoded data.
These, however, are regularization methods of the objective function, and focused on only the 
distribution of the encoded data as the initial value of decoder. 

They did not give the clustering policy based on the decoder structure, 
namely, the reconstruction process of the data. From dynamical system point of view, one decoder of RNN-AE corresponds to a single dynamics in the space of latent representation. Hence, it is natural to equip RNN-AE with multiple decoders to implement multiple dynamics. Such an extension of RNN-AE, however, has yet to be proposed in related works to the best of our knowledge.
It can also possibly be incorporated into the framework of VAE by treating the output of the RNN encoder as a latent random variable \cite{Diederik}.

\section{Recurrent Neural Network and Dynamical System \label{AE}} 
\subsection{Recurrent Neural Network Using Unitary Matrix}
\begin{figure}[h]
\begin{center}
\includegraphics[width=35mm]{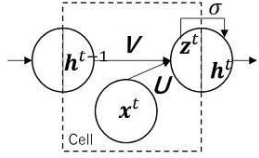}
\end{center}
\caption{RNN Cell}
\label{fig:RNN_Cell}
\end{figure}

RNN is a neural network designed for time series data. The architecture of the main unit is called cell, 
and mathematical expressions are shown in Fig. \ref{fig:RNN_Cell} and Eq. (\ref{eq_dynamical1}).
%,$(\ref{eq_dynamical2})$.

Suppose we are given a time series, 
$$\tX = (\mX^1,\cdots,\mX^n,\cdots,\mX^N), \,\mX^n =  (\vx^1_n,\cdots,\vx^t_n,\cdots,\vx^T_n),$$
$$ \vx^t_n \in \mathbb{R}^D,\, n=1,\cdots,N, \, t=1,\cdots,T,$$
where $D$ denotes data dimension.
RNN, unlike the usual feed-forward neural network, operates the same transform matrix to the 
hidden valuable recursively,
\begin{equation}
\label{eq_dynamical1}
\bm{z}^{t} = \mV \bm{h}^{t-1}+\mU \bm{x}^{t}+ \vb,\,\,\, \bm{h}^t =\sigma(\bm{z}^t),
\end{equation}
where $\sigma(\cdot)$ is an activation function and $\vz^t, \; \vh^t, \; \vb \in \mathbb{R}^L$.
This recursive architecture makes it possible to handle signals with different lengths, although
it is vulnerable to the vanishing gradient problem as with the deep neural network (DNN) \cite{RNN_Elman1,RNN_Elman2}.
Long short-term memory (LSTM) and gated recurrent unit (GRU) are widely known solutions to this problem \cite{LSTM_Felix,LSTM_Hochreiter,GRU_Cho}.
These methods have the extra mechanism called a gate structure to control output scaling and retaining/forgetting of the signal information.  Though this mechanism works effectively in many application fields \cite{Malhotra,Rana}, the architecture of network is relatively complicated.
As an alternative simpler method to solve this problem, the algorithm using a unitary matrix as the transfer matrix $\mV$ was proposed in recent years \cite{Pascanu,Jing1,Wisdom,Arjovsky,Jing2}.
Since the unitary matrix does not change the norm of the variable vector, 
we can avoid the vanishing gradient problem. 
In addition, the network architecture remains unchanged from the original RNN.

In this paper, we focus on the dynamical system aspect of the original RNN.
We employ the unitary matrix type RNN to take advantage of  this dynamical system structure.
However, to implement the above method, we need to find the transform matrix $\mV$ in the space of unitary matrices 
$\mathbb{U}=\{
\mU(L) \in {\rm GL}(L)|\mU(L)^{*}\mU(L) = \mI
\},$
where ${\rm GL}(L)$ is the set of complex-valued general linear matrices with size $L \times L$ and $*$ means the adjoint matrix.
Several methods to find the transform matrix from $\mathbb{U}$ has been reported so far \cite{Pascanu,Jing1,Wisdom,Arjovsky,Jing2}.
Here, we adopt the method proposed by \cite{Jing1}.

\subsection{RNN Autoencoder and Dynamical System}
\begin{figure}[htbp]
\begin{center}
\includegraphics[width=90mm]{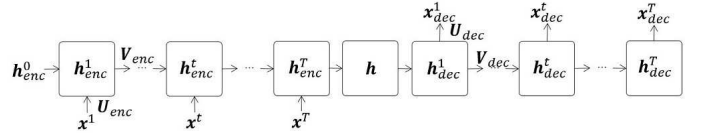}
\end{center}
\caption{Architecture of RNN Autoencoder}
\label{fig:RNN_AE_ARC}
\end{figure}

The architecture of AE using RNN is shown in Fig. \ref{fig:RNN_AE_ARC}. AE is composed of an encoder unit and a decoder unit. The parameters $(\mV_{\it{enc}},\mU_{\it{enc}},\mV_{\it{dec}},\mU_{\it{dec}})$ are trained 
by minimizing  $\|  \mX-\mX_{\it{dec}} \|_F^{2} = \sum_{t=1}^T \|\vx^t-\vx^t_{dec}\|^2$, where $\mX$ is the input data and $\mX_{\it{dec}}$ is the decoded data.

The input data is recovered from only the encoded signal $\vh$ using the matrix $(\mV_{\it{dec}},\mU_{\it{dec}})$, therefore $\vh$ is considered as the essential information of the input signal.
%On the other hand, 
When focusing on the transformation of the hidden variable, this recursive operation has the same structure of a discrete dynamical system expression as described in the following equation:
\begin{equation}
\bm{h}^t = f(\bm{h}^{t-1}),
\end{equation}
where $f$ is given by Eq. (\ref{eq_dynamical1}).
From this point of view, we can understand that RNN describes the universal dynamical system
structure which is common to the all input signals by the reconstruction process in Fig. \ref{fig:RNN_AE_ARC}.
%The dynamical system describes the universal rule in the data  hence we can regard the RNN as the algorithm which drive the universal dynamical rule among the data.
%We take advantage of this property to extract the time series data.
% fig difinition of dynamical system

\section{Derivation of Multi-Decoder RNN AE Algorithm \label{Derivation}}
\begin{figure}[htbp]
\begin{center}
\includegraphics[width=85mm]{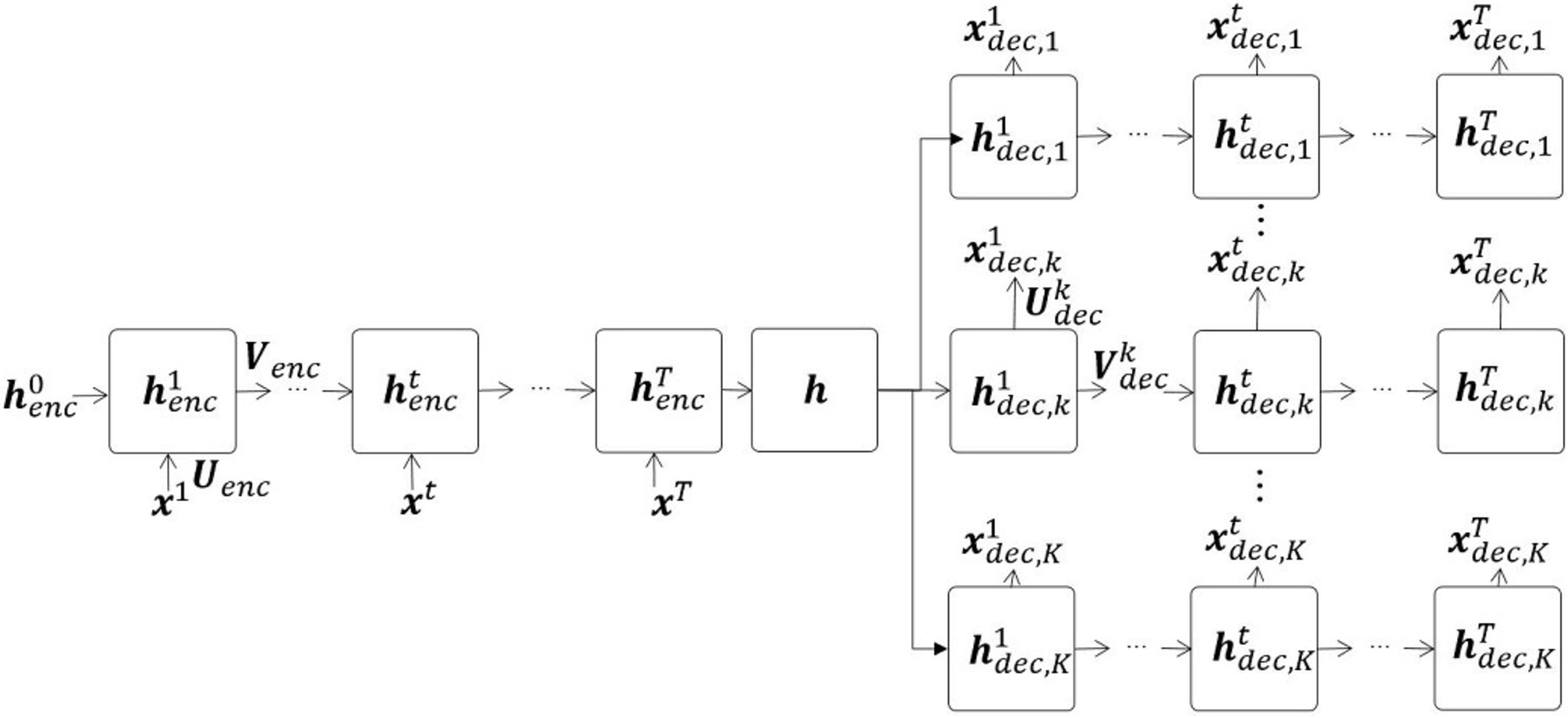}
\caption{Architecture of MDRA}
\label{Multi-decoder RNN AE}
\end{center}
\end{figure}

In this section, we will give the architecture of  the Multi-Decoder RNN AE (MDRA) and its learning algorithm.
As we discussed in the previous section, RNN can extract the dynamical system characteristics of the time series.
In the case of the original RNN, the model expresses just one dynamical system, hence all input data are recovered from the encoded result $\vh$ by the same recovery rule. Therefore $\vh$ is usually 
used as the feature value of the input data.
%This means that all input data are supposed  to obey the same transformation manne
%The architecture of RNN-AE is shown in Fig
%The original data $\bm{x}$ is recovered from the encoded data $\bm{h}$ which is the initial value of the decoder, 
%therefore we consider that the futures of the input data are summarized in $\bm{h}$.
%Whereas, 
In contrast, in this paper, we focus on the transformation rule itself.  
For this purpose, we propose MDRA which has multiple decoders to extract various dynamical system features.
The architecture of MDRA is shown in Fig. \ref{Multi-decoder RNN AE}. 
Let us put $\tW^k_{\it{dec}} = (\mV^k_{\it{dec}},\mU^k_{\it{dec}})$ for $k=1,\cdots,K, \, \tW_{\it{enc}} = (\mV_{\it{enc}},\mU_{\it{enc}})$, and $\tW = (\tW_{\it{enc}},\tW^1_{\it{dec}},\cdots, \tW^K_{\it{dec}})$.
We will derive the learning algorithm to optimize the whole set of parameters $\tW$ in the following section.

\subsection{Decomposition of Free Energy}
We applied a clustering method to derive the learning algorithm of MDRA. 
Many clustering algorithms have been proposed: here we employ the variational Bayes (VB) method, because the VB method enables us to adjust the number of clusters by tuning the hyperparameters of a prior distribution\cite{kaji,Adrian}. 
We first define free energy, which is negative log-marginal-likelihood, by the following equation,
%\begin{equation}
\begin{align}
\label{FreeEnergy}
F_{\tX}(\tW) = -\log \int \int &\left\{ \prod_{n=1}^N \sum_{\vy_n} p_{\tW}(\mX^n|\vy_n, \vh_n,\beta ) p(\vy_n|\bm{\alpha}) \right\} \nonumber \\
&\cdot p( \bm{\alpha})p(\beta)  d\bm{\alpha} d\beta,
\end{align}
%\end{equation}
where $\tX$ is data tensor defined in Section \ref{AE} and $\tW$ is parameter tensor of MDRA defined above.
$\mY=(\vy_1,\vy_2,\cdots,\vy_N)$ is the set of latent variables each of which means an allocation for a decoder. 
That is, $\vy_n = (y_{n1}, \cdots, y_{nK})^{\rm T}\in \mathbb{R}^K$, where $y_{nk}=1$ if $\mX^n$ is allocated to the $k$-th decoder and otherwise $y_{nk}=0$.
$p_{\tW}(\mX^n | \vy_n, \vh_n,\beta )$ is the probability density function representation of MDRA parametrized by tensor $\tW$,
$p( \bm{\alpha})$ and $p(\beta)$ are its prior distributions for a probability vector $\bm{\alpha}=(\alpha_1,\cdots,\alpha_K)$ and a precision parameter $\beta>0$.
We applied  the Gaussian mixture model as our probabilistic model. Hence
$p(\bm{\alpha})$ and $p(\beta)$ were given by Dirichlet and gamma distributions respectively which are the conjugate prior distributions of multinomial and Gaussian distributions. 
These specific distributions are given as follows:
\begin{align*}
& p_{\tW}(\tX,\mY,\bm{\alpha},\beta | \mH) = p_{\tW}(\tX|\mY,\mH,\beta)p(\mY|\bm{\alpha})p(\bm{\alpha})p(\beta), \\
%\end{align*}
%\begin{align*}
& p(\bm{\alpha}) = \frac{\Gamma\left(\theta_0 K\right)}{ \Gamma(\theta_0)^K}\prod_{k=1}^K \alpha_k ^{\theta_0-1}, p(\beta) =\frac{\lambda_0^{\nu_0}}{\Gamma(\nu_0)} \beta^{\nu_0 -1} \exp (- \lambda_0 \beta ), \\
%\end{align*}
%\begin{align*}
& p_{\tW}(\tX|\mY,\mH,\beta)=\prod_{n=1}^N p_{\tW}(\mX^n | \vy_n, \vh_n,\beta ), \\
%\end{align*}
%\begin{align*}
& p_{\tW}(\mX^n | \vy_n, \vh_n,\beta ) =   \\
&  \,\,\,\,\prod_{k=1}^K \left\{ \left (  \frac{\beta}{\pi} \right)^{\frac{T_nD}{2}} \exp (-\beta \| \mX^n - g(\vh_n|\tW_{\it{dec}}^k) \|_F^2 ) \right\} ^{y_{nk}},\\
%\end{align*}
%\begin{align*}
& p(\mY|\bm{\alpha})=\prod_{n=1}^N p(\bm{y}_n|\bm{\alpha}),\,\,\,p(\bm{y}_n|\bm{\alpha}) = \prod_{k=1}^K \alpha_k ^{y_{nk}}. 
\end{align*}
Here, $\theta_0>0, \nu_0>0$ and $\lambda_0 >0$ are hyperparameters and $g(\vh_n| \tW_{\it{dec}}^k)=\mX^n_{dec, k}$ denotes decoder mapping of RNN from the encoded $n$-th data $\vh_n$, $\mH = (\vh_1, \cdots, \vh_N)$ and $T_nD$ is the total signal dimension of input signal $\mX^n$ including dimension of input data.
To apply the variational Bayes algorithm, we then derive the upper bound of the free energy by applying Jensen's inequality,
\begin{align}
%\begin{equation}
%\begin{split}
F_{\tX}(\tW) &= 
%-\log \int \int \left\{ \prod_{n=1}^N \sum_{\vy_n} p_{\tW}(\mX^n|\vy_n, \vh_n,\beta )p(\vy_n|\bm{\alpha}) \right \} p( \bm{\alpha})p(\beta)  d\bm{\alpha} d\beta \nonumber \\
%&= 
-\log \mathbb{E}_{\bar{q}} \left[ 
 \frac{ p_{\tW}(\tX|\mY,\mH,\beta )p(\mY|\bm{\alpha}) p( \bm{\alpha})p(\beta)}
 {q(\mY)q(\bm{\alpha})q(\beta)}
\right] \nonumber \\
&\leq 
\KL( q(\mY)q(\bm{\alpha})q(\beta) \| p(\mY,\bm{\alpha},\beta|\tX))
+
F_{\tX}(\tW)  \nonumber \\
&=\KL( q(\mY)q(\bm{\alpha})q(\beta) \| p(\mY|\bm{\alpha}) p( \bm{\alpha})p(\beta)) \nonumber \\
& \,\,\,\, -\sum_{n=1}^N \mathbb{E}_{\bar{q}'} \left[
\log p_{\tW}(\mX^n|\vy_n,\vh_n,\beta ) 
\right] \nonumber \\%\label{eq:vari_free_enrgy} \\
&\equiv \bar{F}_{\tX}(q,\tW), 
%\end{split}
\end{align}
where $\KL(\cdot \| \cdot)$ is the Kullback$-$Leibler divergence and $\mathbb{E}_{\bar{q}}[\cdot]= \mathbb{E}_{q(\mY)q(\bm{\alpha})q(\beta)}[\cdot] , \mathbb{E}_{\bar{q}'}[\cdot] =\mathbb{E}_{q(\bm{y}_n) q(\beta)}[\cdot]$.
The upper bound $\bar{F}_{\tX}(q,\tW)$ is called the variational free energy or (negated) evidence lower bound (ELBO).
The variational free energy is minimized with respect to the variational posterior $q(\mY, \bm{\alpha},\beta)=q(\mY)q(\bm{\alpha})q(\beta)$  using the variational Bayes method under the fixed parameters $\tW$. 
%The variational free energy is minimized using the variational Bayes method under the fixed 
%parameters $\tW$. %and encode result $\bm{z}$.
Furthermore, it is also minimized with respect to the parameters $\tW$ by applying the RNN learning algorithm to the second term of $\bar{F}_{\tX}(q,\tW)$,%Eq. (\ref{eq:vari_free_enrgy}),
%\[
\begin{align}
&-\sum_{n=1}^N \mathbb{E}_{q(\vy_n)q(\beta) } \left[
 \log p_{\tW}(\mX^n|\vy_n,\vh_n,\beta ) 
\right] \propto \nonumber \\
& \,\,\,\,\,\, \,\,\,\,\sum_{n=1}^N  \mathbb{E}_{q(\vy_n)} \left[
  \sum_{k=1}^K y_{nk}
\| \mX^n -g(\vh_n|\tW_{\it{dec}}^k) \|_F^2 
\right] + const..
%& \,\,\,\,\,\, \,\,\,\,\sum_{n=1}^N  \mathbb{E}_{q(\vy_n)q(\beta)} \left[
%  \sum_{k=1}^K y_{nk}
%\beta \| \mX^n -g(\vh_n|\tW_{\it{dec}}^k) \|_F^2 
%\right].
%\]
\end{align}
%We minimize the $\sum_{k=1}^K \left\{ \bar{\beta} \| \bm{x}_n -g(\bm{z}|W) \|^2  \right\}^{\bar{y}_{nk}}$ by applying the RNN algorithm with 
%\[
%\bar{\beta} = E_{q(\beta)}[\beta],\,\,\,\,\,\,\,\bar{y}_{nk} = E_{q(\bm{y}_n)}[y_{nk}]
%\]

\subsection{Minimization of the Variational Free Energy}
\begin{figure}[h]
\begin{center}
\includegraphics[width=90mm]{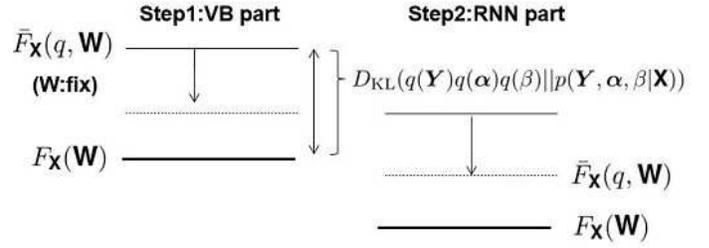}
\vspace{-2mm}
\caption{MDRA algorithm}
\label{fig:MDAE}
\end{center}
\end{figure}

In this section, we derive the variational Bayes algorithm for MDRA to minimize the variational free energy.
We show the outline of the derivation below (for a  detailed derivation, see Appendix \ref{sec:VB_algo} and \ref{sec:RNN_algo} ).
The general formula of the variational Bayes algorithm is given by 
\begin{equation*}
\log q(\mY) = \mathbb{E}_{q(\bm{\alpha},\beta)}[ \log p_{\tW}(\tX, \mY,\mH, \bm{\alpha},\beta)]+const.,
\end{equation*}
\begin{equation*}
\log q(\bm{\alpha},\beta) = \mathbb{E}_{q(\mY)}[ \log p_{\tW}(\tX, \mY,\mH, \bm{\alpha},\beta)]+const..
\end{equation*}
By applying  the above equations to the above probabilistic models (see Appendix \ref{sec:VB_algo}), we obtained the specific algorithm shown in Algorithm \ref{VB-part}. %Appendix \ref{sec:VB_algo}.
\begin{algorithm}[h]
\caption{VB part of MDRA}
\begin{algorithmic} 
\label{VB-part}
\STATE{\bf{Input:} $\tX$: \rm{set of input signals}}
\STATE{\bf{Output:} $\mR$: \rm{allocation weights}} 
%\STATE {Set hyperparameters $\theta_0,\nu_0,\lambda_0$ and the number of iterations $I$ }
%\REPEAT 
\FOR{ $i\leftarrow 0$ to $I$}
\STATE{
\textit{VB E-step}:
\begin{align*}
&\log \rho_{nk} \!=\! \psi(\bar{\theta}_k)\! -\! \psi \! \left( \sum_{k=1}^K \bar{\theta}_k \!\! \right)\!\! -\!\! \|\! \mX^n\! \!-\! g(\vh_n|\tW_{\it{dec}}^k) \|_F^2 \bar{\nu} \bar{\lambda} ^{-1} \\
&\,\,\,\,\, +\! \frac{T_nD}{2}(\psi(\bar{\nu})\!-\! \log \bar{\lambda})\! -\! \frac{T_nD}{2}\log \pi,\,\,
r_{nk} \!=\! \frac{\rho_{nk}}{\sum_{k=1}^K \rho_{nk}}
\end{align*}
} 
\STATE{
\textit{VB M-step}:
\begin{align*}
&N_k = \sum_{n=1}^N r_{nk},\,\,\,\,\bar{\theta}_k = \theta_0 + N_k, \,\,\, \bar{\nu} = \nu_0 + \frac{1}{2} \sum_{n=1}^N T_nD& \\
&\bar{\lambda} = \lambda_ 0 + \sum_{k=1}^K \sum_{n=1}^N r_{nk} \| \mX^n - g(\vh_n|\tW_{\it{dec}}^k) \|_F^2
%&\bar{\nu} = \nu_0 + \frac{1}{2} \sum_{n=1}^N T_nD
\end{align*}
} 
\ENDFOR
\end{algorithmic}
\end{algorithm}
Then we minimize the following weighted reconstruction error using RNN algorithm:
\begin{flalign}
\sum_{n=1}^N  \sum_{k=1}^K \left\{   r_{nk} \| \mX^n -g(\vh_n|\tW_{\it{dec}}^k) \|_F^2  \right\} ,
\end{flalign}
%\begin{flalign*}
%&  \mathbb{E}_{q(\bm{y}_n)}\!\! \left[
% \sum_{n=1}^N \! \sum_{k=1}^K \! y_{nk}
%\beta \| \mX^n \!\! -\!\!g(\vh|\tW_{\it{dec}}) \|^2\! 
%\right] 
%\!\!\! =\!\!
%-\mathbb{E}_{q(\bm{y}_n)} \!\!\! \left[ \! 
%\log \! \left\{ \prod_{k=1}^K \!\! \left \{ \!\! \left(  \frac{\beta}{\pi}  \right)^{\frac{L_nD}{2}}\!\!\!\!\!\! 
%{{e}}^{-\beta \| \mX^n \!-\! g(\vh_n|\tW_{\it{dec}})  \|^2} \right \} ^{y_{nk}} \!\!  \right\} \!\! \right]\\
%&= \sum_{k=1}^K r_{nk} \left\{ \frac{L_nD}{2} (\log \beta-\log \pi) -\beta \| \mX^n - g(\vh_n|\tW_{\it{dec}})  \|^2 \right\},
%\end{flalign*}
where $r_{nk}=\mathbb{E}_{q(\vy_n)}[y_{nk}]$ as detailed in Appendix \ref{sec:RNN_algo}. 
We denote $\mR=(\vr_1, \cdots, \vr_N)$, where $\vr_n = (r_{n1}, \cdots, r_{nK})^{\rm T} \in \mathbb{R}^K$.
From the above discussion, we finally obtained the following Algorithm \ref{RNN-part}.
We apply these two algorithms iteratively to minimize $\bar{F}_{\tX}(q,\tW)$.
\begin{algorithm}[h]
\caption{MDRA}
\begin{algorithmic} 
\label{RNN-part}
\STATE{\bf{Input:} $\tX$: \rm{set of input signals}}
\STATE{\bf{Output:} $\tW$: \rm{weight tensors,} $\mR$: \rm{allocation weights,} $\mH$: \rm{encoded signals} }
\STATE {Set hyperparameters $\theta_0,\nu_0,\lambda_0$ and the initial value of $\tW$ randomly.}
\REPEAT 
\STATE{
Calculate $\tW$ that minimizes the following value by RNN algorithm:
        \[
         \sum_{n=1}^N  \sum_{k=1}^K \left\{   r_{nk} \| \mX^n -g(\vh_n|\tW_{\it{dec}}^k) \|_F^2  \right\}.
        \] 
} 
\STATE{Calculate $\mR=(r_{nk})$ by the algorithm \textit{VB part of MDRA}\,\,(Algorithm \ref{VB-part}).  }
%\UNTIL{ $\sum_{n=1}^N  \sum_{k=1}^K \left\{   y_{nk} \| \bm{x}_n -g(\bm{h}_n|W) \|^2  \right\}\,\,<$\,\,Threshold  }
\UNTIL{ the difference of variational free energy $\bar{F}_{\tX}(q,\tW)\,\,<$\,\,Threshold  }
\end{algorithmic}
\end{algorithm}
Fig. \ref{fig:MDAE} describes the relation of the VB  and RNN steps of MDRA algorithm. 

\section{Experiments \label{Exp}}
\subsection{ Periodic Signals \label{Exp1}}
\begin{figure}[h]
\begin{center}
\includegraphics[width=90mm,height=25mm]{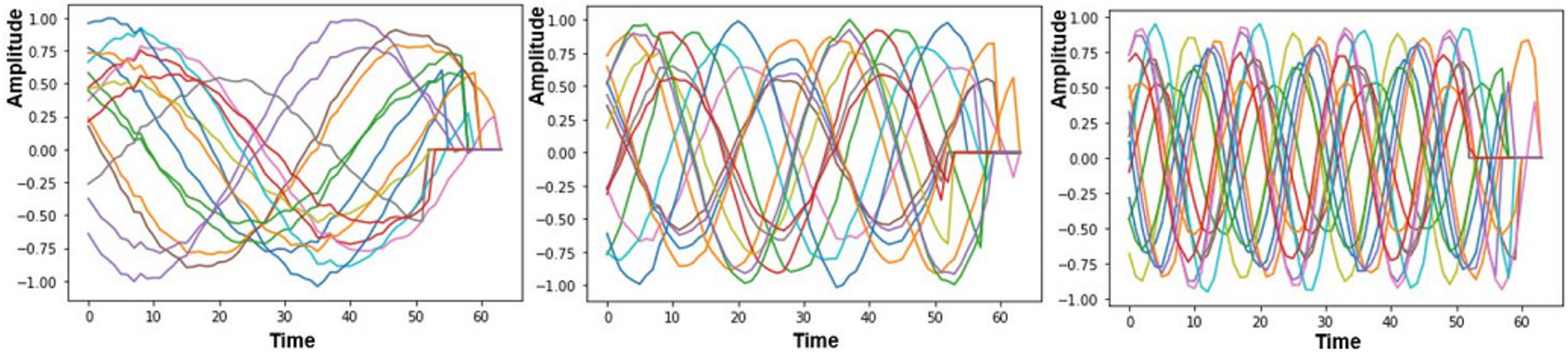}
\caption{Examples of periodic signals}
\label{fig:INPUT_TOY}
\end{center}
\end{figure}

We first examined the basic performance of our algorithm using periodic signals.
Periodic signals are typical time series signals expressed by dynamical systems.
Input signals have 2, 4, and 8 periods respectively in 64 steps. 
Each signal is added a phase shift (maximum one period), amplitude variation (from 50\% to 100\% of the maximum amplitude), 
additional noise (maximum 2\% of maximum amplitude) and signal length variation (maximum 80\% of the maximum signal length).
Examples of input data are illustrated in Fig. \ref{fig:INPUT_TOY}. 

\begin{figure}[h]
\begin{center}
\includegraphics[width=80mm]{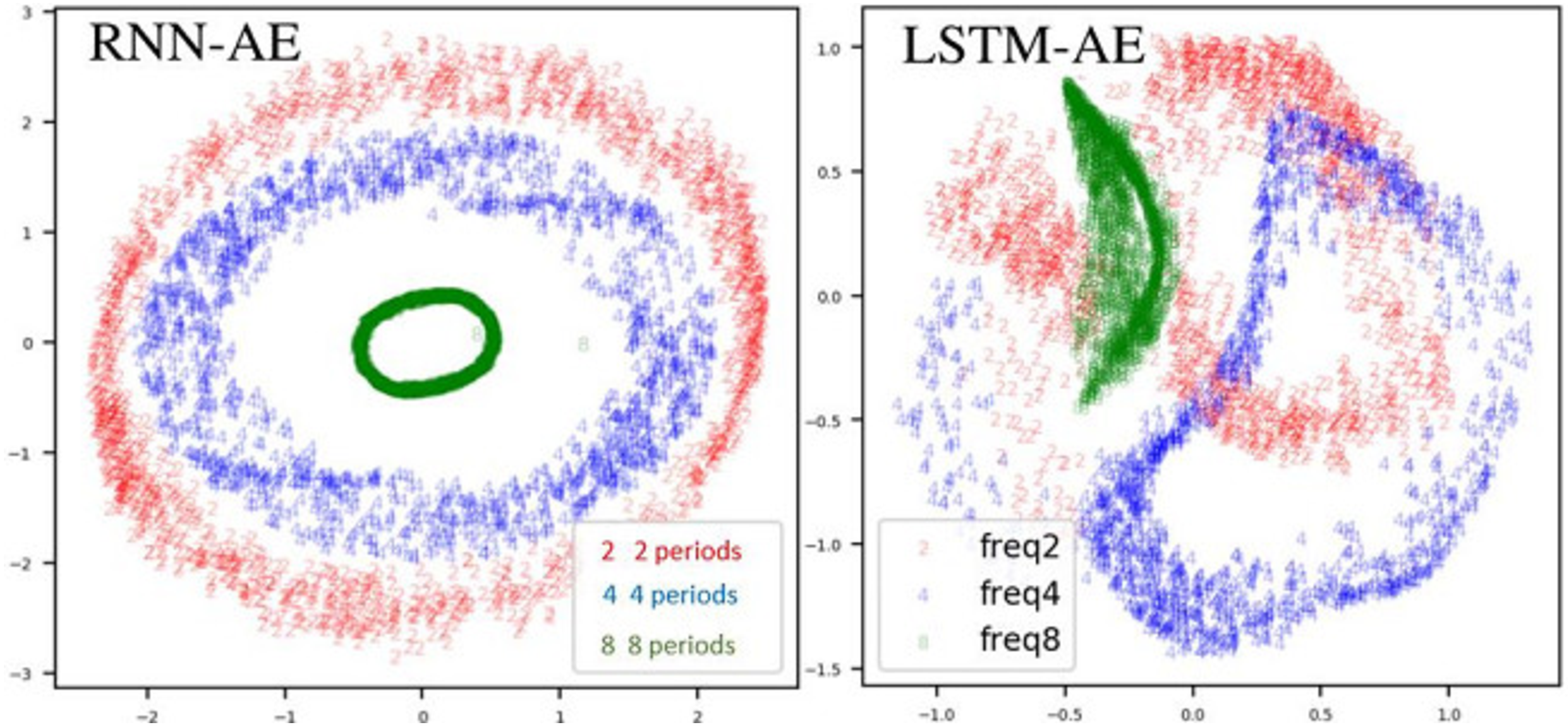}
\caption{Visualization of features extracted by RNN-AE: left, LSTM-AE: right}
\label{fig:RNN_AE_RES}
\end{center}
\end{figure}

\begin{figure}[h]
\begin{center}
\includegraphics[width=80mm]{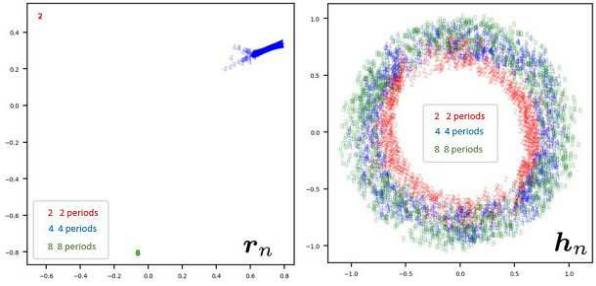}
\caption{Visualization of features extracted by MDRA (left: $\vr_n$, right: $\vh_n$)}
\label{fig:MDAE_RES}
\end{center}
\end{figure}

We compared LSTM-AE and RNN-AE to MDRA on its feature extraction performance using the above periodic signals.
Fig. \ref{fig:RNN_AE_RES} and Fig. \ref{fig:MDAE_RES} show the results of LSTM-AE, RNN-AE and MDRA, respectively.
We set the same dimension of hidden variable $\vh_n$ in all algorithms. Note here that RNN-AE and MDRA use a complex-valued hidden variable while LSTM-AE uses real-valued one.
%LSTM-AE uses real number as the hidden variable.
Therefore LSTM-AE has twice the hidden variable dimension of RNN-AE and MDRA.  
The parameter setting is listed in Table \ref{tbl:Priodic} in Appendix \ref{ParamSet}.

We used multi-dimensional scaling (MDS) as the dimension reduction method to visualize the distributions of features in Fig. \ref{fig:RNN_AE_RES} and Fig. \ref{fig:MDAE_RES}. 

Fig. \ref{fig:RNN_AE_RES} shows the distribution of the encoded data $\vh_n$ which is the initial value of the decoder unit in Fig. \ref{fig:RNN_AE_ARC}.

We found that RNN-AE can separate the input data into three regions corresponding to each frequency (Fig. \ref{fig:RNN_AE_RES}:right).
%However each frequency region is distributed widely, therefore some part of the region overlap each other. 
However distribution on the hidden variable of LSTM-AE has complicated shape, each frequency overlapped 
each other.  
We guess this result was caused by the complex architecture of LSTM cell.

Fig. \ref{fig:MDAE_RES} shows the distributions of the encoded data $\vh_n$ and the clustering allocation weight $\vr_n$ extracted by MDRA.
The distribution of $\vr_n$ shown in the left figure of Fig. \ref{fig:MDAE_RES} is completely separated into each frequency component without overlap. The distribution of $\vh_n$ was given as the initial value of the corresponding decoder. 
This result shows that the distribution of $\vr_n$ as the feature extraction has robustness for phase shift, amplitude and signal length variation.
%We also show that MDRA can boost the classification accuracy using an actual driving data in the next section. 

\subsection{ Complex Periodic Signals \label{Exp1_2}}
\begin{figure}[h]
\begin{center}
\includegraphics[width=90mm]{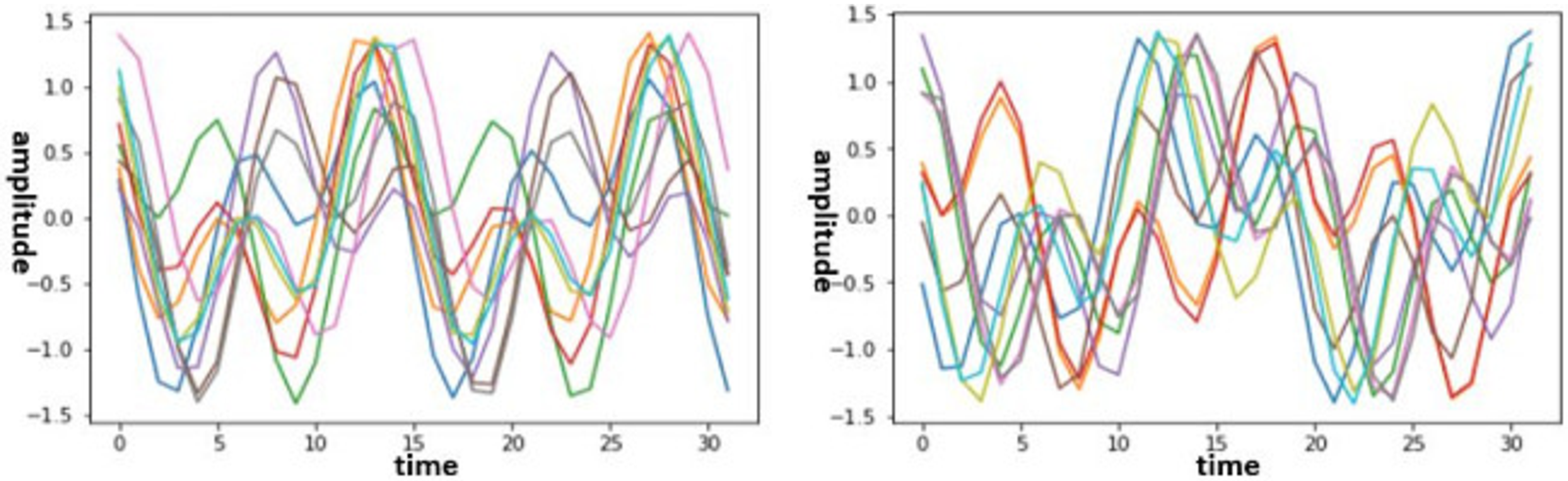}
\caption{Examples of complex periodic signals}
\label{fig:compositive_sig}
\end{center}
\end{figure}

Next we applied our algorithm to more complicated signals. The input signals were all length $32$ steps and created by the following steps.
\begin{enumerate}
  \item Give $\theta_i \in [0,2\pi], i=1,2$ randomly. 
  \item Set $\vh_n^{0}  =(e^{i\theta_1},e^{i\theta_2})$.%=(h_0^1,h_0^2) =(e^{i\theta_1},e^{i\theta_2})
  \item Create $\vh_{n}^t \in \mathbb{C}^2$ by the rule $\vh_{n}^{t+1} =\begin{pmatrix}
e^{i\omega_1} & 0 \\
0 & e^{i\omega_2} \\
\end{pmatrix}
\vh_n^{t},\,\,\,\,(t=1,2,\cdots,31)$.
  \item Obtain the signal $x_n^t$ by projecting $\vh_n^t$ to the vector $(1,1,1,1)$ as the real value vector  $\vh_{n}^t \in \mathbb{R}^4$
\end{enumerate}
We created two types of signals (5000 for each type) with A:$(\omega_1,\omega_2)=(55.0,20.0)$ and B:$(\omega_1,\omega_2)=(50.0,25.0)$, respectively. We found that it is not very easy to separate the two types of signals from Fig. \ref{fig:compositive_sig} visually. 

\begin{figure}[h]
\begin{center}
\includegraphics[width=80mm]{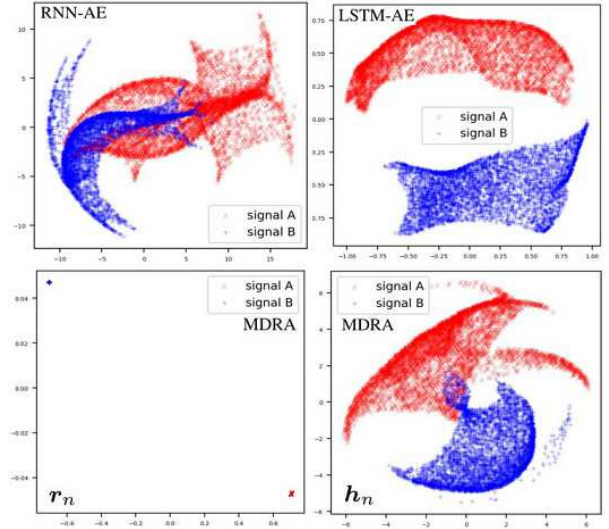}
\caption{Visualization of features extracted by  RNN-AE: top left, LSTM-AE: top right and MDRA: bottom left and right (complex periodic signals)}
\label{fig:COMPOSITE_LSTM_RES}
\end{center}
\end{figure}

Fig. \ref{fig:COMPOSITE_LSTM_RES} shows the result of each algorithm applied to the complex periodic signals. We used the same hidden variable dimension for all algorithms.
Further information on the parameters are listed in Table \ref{tbl:cpxPeridic} in Appendix \ref{ParamSet}.   
Unlike the experiment \ref{Exp1}, although RNN-AE could not separate the two types of signals completely, LSTM-AE
was able to separate them. Furthermore $\vr_n$ of MDRA classified the signals based on the periodicity without any influence from the phase shift. The phase shift was expressed by $\vh_n$ similarly to the experiment \ref{Exp1}.

\begin{figure}[h]
\begin{center}
\includegraphics[width=80mm]{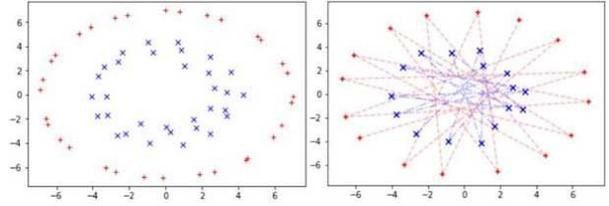}
\caption{Trajectory of hidden variable of MDRA (signal A: red, signal B: blue)}
\label{fig:COMPOSITE_MDRA_RES}
\end{center}
\end{figure}

In this experiment, the MDRA estimated the number of clusters and data ratios correctly in spite of the setting of the number of decoders $K=5$. The distribution ratios calculated from $\vr_n$ for $2$ major clusters were $49.8$\% and $49.0$\%.
Fig. \ref{fig:COMPOSITE_MDRA_RES}:left is the MDS expression of hidden variable trajectories. 
Fig. \ref{fig:COMPOSITE_MDRA_RES}:right shows the first eight signals with successive data connected by lines. We found that these two types of signals were completely expressed as different periodic signals in the hidden space.  

%\newpage
\subsection{Experiment of Real Driving Data\label{Exp2}}

We applied our algorithm to a real driving data clustering problem. 
We use the driving data consisting of speed, acceleration, braking and steering angle signals.\footnote{ This data was created by HQL (Research Institute of Human Engineering for Quality Life:
https:\slash\slash{} www.hql.jp\slash{} howhql\slash{} spirit.html).}
The input signal was about 1 minute  differential data, which was cut out from the original data by a sliding window.\footnote{We use only the data of which the maximum acceleration difference is more than a certain threshold.}
The detailed information of the input data is shown in Table \ref{fg:DrivSetting}.
\begin{table}[h]
\centering
\caption{Driving data clustering}
%\label{tab:my-table}
\begin{tabular}{|c|c|c|c|c|c|}
\hline
\#Training & Signal length & Sampling pitch & Slide \\ \hline
4644    & 512           & 0.1 sec.       & 8     \\ \hline
\end{tabular}
\label{fg:DrivSetting}
\end{table}

The feature extraction results by MDRA are shown in Fig. \ref{fig:DRIVE_RES}. 
The parameter setting of this experiment is listed in Table \ref{tbl:Route} in Appendix \ref{ParamSet}. 
The left figure is the route clustering result based on the driving behavior by the MDRA ($K=10$).  
%Fig. \ref{fig:DRIVE_RES} 
This figure shows the actual trajectory of a driven car, each point of which is colored by RGB based on $3$ dimensional representation of $\vr_n$ given by the MDS. 
The right figures (No.1-No.4) show the typical driving behavior extracted from the major clusters.
Blue, green, orange and red lines are speed, acceleration, brake and steering angle, respectively.  
From these results, the interpreted driving feature of each cluster and its ratio are as follows:
\begin{itemize}
	\item No.1: moderate acceleration 14.0\%
      \item No.2: stable travel (high speed) 7.5\% 
      \item No.3: moderate deceleration 3.5\%
	\item No.4: stable travel (middle speed) 13.7\%
\end{itemize}
In addition, we can extract the complicated driving operation such as No.5 by choosing the low ratio data point which is significantly different from the surrounding data points. 

\begin{figure}[h]
\begin{center}
\includegraphics[width=90mm]{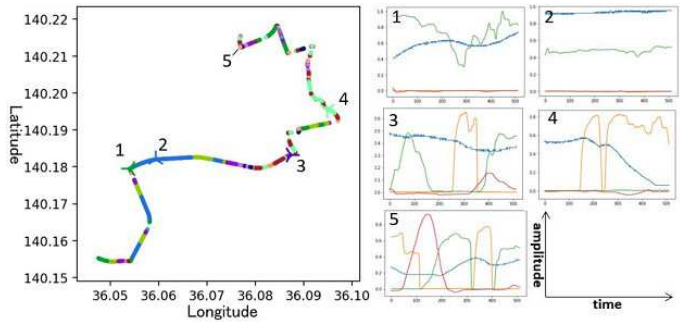}
\caption{Clustering result of driving data ($K=10$) }
\label{fig:DRIVE_RES}
\end{center}
\end{figure}

Although we showed the result in the case of $K=10$ here, we can adjust the clustering size by changing $K$ and the hyparparameters.
%The clustering result ($K=3$) is shown in Fig.\ref{fig:DRIVE_RES_K3}.

%\begin{figure}[h]
%\begin{center}
%\includegraphics[width=50mm]{figures2/route_K3.EPS}
%\caption{Classification result of driving data($K=3$) }
%\label{fig:DRIVE_RES_K3}
%\end{center}
%\end{figure}

\section{Discussion}
We verified  the feature extraction performance of the MDRA using actual time series data.
In Sections \ref{Exp1} and \ref{Exp1_2}, we saw that 
MDRA algorithm can achieve more stable clustering than LSTM-AE and RNN-AE by using decoder weight $\vr_n$ for periodic and complex periodic data.
%in cases of different signal lengths, variation of amplitude, additional noise and complex signals. 
In addition, we also showed that MDRA has the function to reduce the unnecessary clusters using the property of the variational Bayes method. 
In Section \ref{Exp2}, we confirmed that above variational Bayes property provides the flexible clustering and uncommon data extraction using the actual driving data. 
There are a lot of research on the variational Bayes method\cite{nakajima}, therefore we can apply these algorithms and knowledges to improve the performance of MDRA.
Especially the phase transition phenomenon of the variational Bayes learning method, depending on the hyperparameters, has been reported in \cite{watanabe}．The hyperparameter setting of the prior distribution has a great effect on the clustering result.

\section{Conclusion\label{Summary}}
In this paper, we proposed a new clustering algorithm, MDRA, which can extract features of time series data based on the data generating process expressed by decoders.
We conducted experiments using periodic signals and actual driving data to verify the advantages of MDRA.
The results show that our algorithm has not only robustness for the phase shift, amplitude, signal length variation, and signal synthesis but also flexibility on the clustering performance.
We intend to undertake a detailed study of the relation between the feature
extraction performance and hyperparameter setting of the prior distributions in the future.

%\section{Future work\label{Future}}
%The phase transition phenomenon of the variational Bayes learning method, depending on the hyperparameters, has been reported in \cite{watanabe}．The hyperparameter setting of the prior distribution has a great effect on the clustering result and classification performance.
%We intend to undertake a detailed study of the relation between the feature
%extraction performance and hyperparameter setting of the prior distributions in the future.

%\newpage

%\section*{Acknowledgment}
%The preferred spelling of the word ``acknowledgment'' in America is without 
%an ``e'' after the ``g''. Avoid the stilted expression ``one of us (R. B. 
%G.) thanks $\ldots$''. Instead, try ``R. B. G. thanks$\ldots$''. Put sponsor 
%acknowledgments in the unnumbered footnote on the first page.

%\section*{References}
%\input{Sec/reference.tex}
\bibliographystyle{myIEEEbib}
\bibliography{reference}

%\newpage
%\newpage
\appendix
%\def\thesection{補遺\Alph{section}}
%\section{Derivation of Algorithm}
%\label{sec:DV_algo} 
%\subsection{Minimization algorithm of the variational free energy}
\subsection{Minimization of Variational Free Energy with Respect to the Variational Posterior for the Fixed RNN Parameter}
%\subsection{Minimization of variational free energy with respect to $q(\mY, \bm{\alpha}, \beta)$ for fixed $\tW$}
\label{sec:VB_algo}
%\label{Ape1}
%\begin{flushleft}
%{\bf \textit{Minimization of variational free energy with respect to $q(\mY, \bm{\alpha}, \beta)$ for fixed $\tW$}:}\\
%\end{flushleft}
Initially, we suppose that the posterior is expressed by $q(\mY, \bm{\alpha}, \beta )=q(\mY) q(\bm{\alpha}, \beta)$.
Then
\begin{equation*}
\begin{split}
\log q(\mY) \!\! &=\! \mathbb{E}_{q(\bm{\alpha},\beta)}[ \log p_{\tW}(\tX, \mY, \mH,\bm{\alpha},\beta)]+const. \\
&=\!\mathbb{E}_{q(\bm{\alpha},\beta)}[ \log p_{\tW}(\tX|\mY,\mH, \beta) ] \!+\! \mathbb{E}_{q(\bm{\alpha},\beta)}[ \log p(\mY|\bm{\alpha}) ] \\
&\quad + \! \mathbb{E}_{q(\bm{\alpha},\beta)}[ \log p(\bm{\alpha}) ]+\mathbb{E}_{q(\bm{\alpha},\beta)}[ \log p(\beta) ]+const.\\
&=\! \mathbb{E}_{q(\bm{\alpha})}[ \log p(\mY|\bm{\alpha}) ] + \mathbb{E}_{q(\bm{\alpha})}[ \log p(\bm{\alpha}) ]\\
&\quad \!+\!\! \mathbb{E}_{q(\!\beta \!)}[ \log p_{\tW}(\tX|\mY, \mH,\beta) ] \!\!+ \!\! \mathbb{E}_{q(\! \beta\!)}[ \log p(\beta) ]\!\!+\!const..
\end{split}
\end{equation*}
In addition, 
\begin{align*}
\mathbb{E}_{q(\bm{\alpha})}[\log p(\mY | \bm{\alpha} )] &= \mathbb{E}_{q(\bm{\alpha})}\left [\log \prod_{n=1}^N \prod_{k=1}^K \alpha_k ^{y_{nk}} \right]\\&= \sum_{n=1}^N \sum_{k=1}^K y_{nk} \mathbb{E}_{q(\bm{\alpha})}   \left [\log \alpha_k  \right],
\end{align*}
\begin{align*}
& \mathbb{E}_{q(\beta)} \left[  \log p_{\tW}( \tX| \mY,\mH,\beta)   \right] \\
& = \sum_{n=1}^N \sum_{k=1}^K y_{nk} \mathbb{E}_{q(\beta)} \!
\left[
-\beta \| \mX^n - g(\vh_n | \tW_{\it{dec}}^k) \|_F^2 
 \!+\!  \frac{T_nD}{2}\log \frac{\beta}{\pi} \! \right],  
\end{align*}
where $T_nD$ means total signal dimension. Therefore, we obtain 
\[
\log q(\mY) = \sum_{n=1}^N \sum_{k=1}^K y_{nk} \log \rho_{nk} + const..
\]
We here put 
\begin{align}
\log \rho_{nk} = \mathbb{E}_{q(\bm{\alpha})}   \left [\log \alpha_k  \right] 
+  \mathbb{E}_{q(\!\beta )} 
\left[ G\right], \label{eq:rho_nk}
\end{align}
where 
$G=G' \!+\!   \frac{T_nD}{2}(\log \beta -\log \pi), G'=-\beta \| \mX^n \!-\! g(\vh_n | \tW_{\it{dec}}^k) \|_F^2$.
Hence 
$
q(\mY) \propto \prod_{n=1}^N \prod_{k=1}^K \rho_{nk}^{y_{nk}}
$, 
%$\sum_{k=1}^N \rho_{nk} = 1$ is required, so by putting 
by putting  
$
r_{nk} = \frac{\rho_{nk}}{\sum_{k=1}^K \rho_{nk}},
$
we obtain
%\noindent\fbox{\parbox{\linewidth-2\fboxsep-2\fboxrule}{%
\begin{flalign*}
q(\mY) = \prod_{n=1}^N \prod_{k=1}^K  r_{nk}^{y_{nk}}.
\end{flalign*}
%}}
Next we calculate $\log q(\bm{\alpha}, \beta)$,
\begin{equation*}
\begin{split}
\log q(\bm{\alpha},\beta) &= \mathbb{E}_{q(\mY)}[ \log p_{\tW}(\tX, \mY,\mH, \bm{\alpha},\beta)]+const. \\
%&=E_{q(\bm{y})}[ \log p_{W,z}(\bm{X}|\bm{Y}, \beta) ] + E_{q(\bm{y})}[ \log p(\bm{Y}|\alpha) ] \\
%&\quad +E_{q(\bm{y})}[ \log p(\alpha) ]+E_{q(\bm{y})}[ \log p(\beta) ]+const.\\
&=\mathbb{E}_{q(\mY)}[ \log p(\mY|\bm{\alpha}) ] + \log p(\bm{\alpha}) \\
&\!\!\quad \!+\! \mathbb{E}_{q(\! \mY\! )\!}[ \log p_{\tW}(  \tX|\mY,\mH, \beta) \!] \!+\! \log p(\beta) \!+\!const..
\end{split}
\end{equation*}
Above equation  can be divided into the two terms including $\bm{\alpha}$ and $\beta$ respectively, 
\begin{equation*}
\begin{split}
&\log q(\bm{\alpha}) \propto \mathbb{E}_{q(\mY)} [\log p(\mY|\bm{\alpha})]\! + \! \log p(\bm{\alpha})\! +\! const.\\
%&=  \mathbb{E}_{q(\bm{y})} \left[
%\sum_{n=1}^N \sum_{k=1}^K y_{nk} \log \alpha_k
%\right]
%+\log  C_0(\alpha_0) + (\alpha_0-1)\sum_{k=1}^K \log \alpha_k +const.\\
&=  \sum_{n=1}^N \sum_{k=1}^K \log \alpha_k  \mathbb{E}_{q(\vy_n)} \left[
 y_{nk} 
\right]
 + (\theta_0-1)\sum_{k=1}^K \log \alpha_k +const..
\end{split}
\end{equation*}
Substituting 
$
\mathbb{E}_{q(\vy_n)}[y_{nk}] = 1\cdot q(y_{nk}=1)+0\cdot q(y_{nk} = 0 ) = q(y_{nk} =1 ) = r_{nk}
$
to the above equation, we obtain
%\noindent\fbox{\parbox{\linewidth-2\fboxsep-2\fboxrule}{%
\begin{flalign*}
\log q(\bm{\alpha}) = \sum_{n=1}^N \sum_{k=1}^K \log \alpha_k r_{nk} + (\theta_0 -1 )\sum_{k=1}^K \log \alpha_k + const. .
\end{flalign*}
%}}
On the other hand,
\begin{align*}
%\begin{split}
&\log q(\beta) = \mathbb{E}_{q(\mY)} [\log p_{\tW}(\tX|\mY,\mH,\beta)] +\log p(\beta) + const.\\
&=\! \sum_{n=1}^N   \sum_{k=1}^K  \left[ E_{q(\vy_n)}  \left[
y_{nk}  \right]  \cdot G \right] +(\nu_0-1)\log \beta + \lambda \beta +const..
%\end{split}
%\! \left\{ \! - \beta \| \mX^n \!-\!  g(\vh_n|\tW_{\it{dec}}^k) \|_F^2 \!+\! \frac{T_nD}{2}\log\frac{\beta}{\pi} \!\right\} \! \right] \\
\end{align*}
%Similarly, 
By applying $\mathbb{E}_{q(\bm{y}_n)}[y_{nk}] = r_{nk} $, we obtain
%\noindent\fbox{\parbox{\linewidth-2\fboxsep-2\fboxrule}{%
\begin{flalign*}
& \log q(\beta) \\
& \,\,\,\,\,\,\,\,=\sum_{n=1}^N \sum_{k=1}^K \left[ r_{nk} \cdot G \right ]+(\nu_0-1)\log \beta + \lambda \beta +const..
\end{flalign*}
%}}
We finally calculate $\log \rho_{nk}$ in Eq. (\ref{eq:rho_nk}). 
% to obtain $r_{nk}$.
We first calculate $\mathbb{E}_{q(\!\beta )} 
\left[ G\right]$,%$\log q(\bm{\alpha})$ and $\log q(\beta)$,
\begin{equation*}
\begin{split}
\log q(\beta) = \beta  f \!+\! \left (\nu_0 + \frac{1}{2}\sum_{n=1}^N T_nD  -1 \right )\log \beta\! -\!  \lambda \beta \!+\! const., 
%\sum_{n=1}^N  \sum_{k=1}^K \left[ - r_{nk}\left\{  \beta \| \mX^n - g(\vh_n|\tW_{\it{dec}}^k) \|_F^2 + \frac{T_nD}{2}(\log \beta- \log \pi)\right\} \right]\\
%&+ (\nu_0-1)\log \beta - \lambda_0 \beta +const.\\
%&= \beta  \sum_{k=1}^K  \sum_{n=1}^N  - r_{nk}  \| \bm{x}_n - g(\bm{z}_n|W) \|^2 +  \sum_{k=1}^K  \sum_{n=1}^N r_{nk}\frac{T_n}{2}(\log \beta- \log 2\pi)\\
% &+ (\nu_0-1)\log \beta + \lambda_0 \beta +const.\\
% &= \beta  f + \left (\nu_0 + \frac{1}{2}\sum_{n=1}^N T_n  -1 \right )\log \beta - \lambda \beta + \frac{1}{2} \sum_{n=1}^N T_n \log 2\pi+ const.\\
% &= \beta  f + \left (\nu_0 + \frac{1}{2}\sum_{n=1}^N T_nD  -1 \right )\log \beta - \lambda \beta + const.,
\end{split}
\end{equation*}
where we put 
$
f = \sum_{k=1}^K \sum_{n=1}^N  - r_{nk}  \| \mX^n - g(\vh_n|\tW_{\it{dec}}^k) \|_F^2.
$
In addition, putting $\bar{\lambda} = \lambda_0 - f$,\,\,$\bar{\nu}=\nu_0+\frac{1}{2}\sum_{n=1}^N T_nD$, 
\begin{equation*}
\begin{split}
q(\beta)\! &=\!  {e}^{ \beta  f} \!  \beta^{\nu_0+\frac{1}{2}\sum_{n=1}^N T_nD-1} {e}^{-\!  \lambda_0 \beta}\! \cdot \! const.
%&=   e^{-(\lambda_0-f ) \beta } \beta^{\nu_0+\frac{1}{2}\sum_{n=1}^N T_n-1}\cdot const.\\
\!\!=\!\! {e}^{-\bar{\lambda} \beta } \beta^{\bar{\nu}-1}\!\cdot\! const. \\
&=\frac{\bar{\lambda} ^{\bar{\nu}}}{\Gamma(\bar{\nu})}  \beta^{\bar{\nu}-1} {e}^{-\bar{\lambda} \beta} = Gamma(\beta|\bar{\nu},\bar{\lambda}).
\end{split}
\end{equation*}
By using the expectations of $\beta$ and $ \log \beta $ by gamma distribution $\mathbb{E}_{q(\beta)}[\beta]=\nu \lambda^{-1}, \mathbb{E}_{q(\beta)}[\log \beta] = \psi(\nu) - \log \lambda$ (Appendix \ref{DeriveEqlog}), we obtain 
\begin{flalign*}
\mathbb{E}_{q(\beta)} \left[
%-\beta \| \mX^n - g(\vh_n | \tW_{\it{dec}}^k) \|_F^2 + \frac{T_nD}{2}(\log \beta- \log \pi)  
G
\right]  
%& \,\,\,\,\,= - \| \bm{x}_n - g(\bm{z}_n|W) \|^2 E_{q(\beta)}
%\left[
%\beta 
%\right] + \frac{T_n}{2} E _{q(\beta)} [\log \beta]- \frac{T_n}{2}\log 2\pi    \\
&= -\| \mX^n - g(\vh_n|\tW_{\it{dec}}^k) \|_F^2 \bar{\nu} \bar{\lambda} ^{-1} \\
&+ \frac{T_nD}{2}(\psi(\bar{\nu})-\log \bar{\lambda})- \frac{T_nD}{2}\log \pi .
\end{flalign*}
%On the other side, %$E_{q(\alpha)}[\log \alpha_k ]$ is calculated the same way as the general mixture model,
Similarly, $q(\bm{\alpha})$ turns out to be the Dirichlet distribution with parameters $(\bar{\theta}_1, \cdots, \bar{\theta}_K)$, and
$
\mathbb{E}_{q(\bm{\alpha})}[\log \alpha_k ] = \psi(\bar{\theta}_k) -\psi \left( \sum_{k=1}^K \bar{\theta}_k \right)
$
is calculated by the same way in the general mixture model\cite{Attias,Ghahramani,kaji}. Therefore we finally obtain
\begin{equation*}
\begin{split}
\log \rho_{nk} &= \psi(\bar{\theta}_k) -\psi \left( \sum_{k=1}^K \bar{\theta}_k \right) - \| \mX^n - g(\vh_n|\tW_{\it{dec}}^k) \|_F^2 \bar{\nu} \bar{\lambda} ^{-1}\\
&+ \frac{T_nD}{2}(\psi(\bar{\nu})-\log \bar{\lambda}) - \frac{T_nD}{2}\log \pi .
\end{split}
\end{equation*}
From the above results, the following variational Bayes algorithm is derived.
%we obtain the following iterative algorithm:\\

%\begin{algorithm}[h]
%\caption{VB part of MDRA}
%\begin{algorithmic} 
%\label{VB-part}
%\STATE{\bf{Input:} $\tX$: \rm{set of input signals}}
%\STATE{\bf{Output:}$\mR$: \rm{allocation weight}} 
%\STATE {Set hyperparameters $\theta_0,\nu_0,\lambda_0$ and the number of iterations $I$ }
%\REPEAT 
%\FOR{ $i\leftarrow 0$ to $I$}
%\STATE{
%\textit{VB E-step}:
%\begin{flalign*}
%\log \rho_{nk} &= \psi(\bar{\theta}_k) -\psi \left( \sum_{k=1}^K \bar{\theta}_k \right) - \| \mX^n - g(\vh_n|\tW_{\it{dec}}^k) \|_F^2 \bar{\nu} \bar{\lambda} ^{-1}\\
%&+ \frac{T_nD}{2}(\psi(\bar{\nu})-\log \bar{\lambda}) - \frac{T_nD}{2}\log \pi \\
%r_{nk} &= \frac{\rho_{nk}}{\sum_{k=1}^K \rho_{nk}}
%\end{flalign*}
%} 
%\STATE{
%\textit{VB M-step}:
%\begin{align*}
%N_k &= \sum_{n=1}^N r_{nk},\,\,\,\,\bar{\theta}_k = \theta_0 + N_k & \\
%\bar{\lambda} &= \lambda_ 0 + \sum_{k=1}^K \sum_{n=1}^N r_{nk} \| \mX^n - g(\vh_n|\tW_{\it{dec}}^k) \|_F^2,\,\,\,\,
%\bar{\nu} = \nu_0 + \frac{1}{2} \sum_{n=1}^N T_nD
%\end{align*}
%} 
%\ENDFOR
%\end{algorithmic}
%\end{algorithm}

%\subsection{Minimization algorithm of the RNN term (the second term of variational free energy)}
\subsection{Minimization of Variational Free Energy with Respect to the RNN Parameter for the Fixed Variational Posterior}
%\subsection{Minimization of variational free energy with respect to $\tW$ for fixed $q(\mY, \bm{\alpha}, \beta)$}
\label{sec:RNN_algo}
%\begin{flushleft}
%{\bf \textit{Minimization of variational free energy with respect to $\tW$ for fixed $q(\mY, \bm{\alpha}, \beta)$}:}\\
%\end{flushleft}
We minimize 
\[
-\mathbb{E} _{q(\mY)q(\beta)}\left[
\sum _{n=1}^N \log p_{\tW}(\mX^n | \vy_n,\vh_n,\beta)
\right]
\]
to minimize the free energy Eq. (\ref{FreeEnergy}) with respect to $\tW$.
More specifically, we minimize
%\vspace{-5mm}

\begin{flalign*}
& -\sum_{n=1}^N \mathbb{E}_{q(\vy_n)q(\beta)}\left[\log \left\{ \prod_{k=1}^K \left \{ \left(  \frac{\beta}{\pi}  \right)^{\frac{T_nD}{2}} 
e^{G'} \right \} ^{y_{nk}} \right\}\right] \\
& =-\sum_{n=1}^N \mathbb{E}_{q(\vy_n)q(\beta)}\left[\sum_{k=1}^K y_{nk} \left\{\frac{T_nD}{2}(\log \beta -\log \pi)G'\right\} \right]\\
%& =-\sum_{n=1}^N \sum_{k=1}^K \mathbb{E}_{q(\vy_n)}[y_{nk}] \left\{\frac{T_nD}{2}(\mathbb{E}_{q(\beta)}[\log \beta] -\log \pi) -\mathbb{E}_{q(\beta)}[\beta] \| \mX^n - g(\vh_n|\tW_{\it{dec}}^k)  \|_F^2\right\} \\
%& =\sum_{n=1}^N \sum_{k=1}^K r_{nk} \left\{ \mathbb{E}_{q(\beta)}[\beta] \| \mX^n - g(\vh_n|\tW_{\it{dec}}^k)  \|_F^2 -\frac{T_nD}{2}(\mathbb{E}_{q(\beta)}[\log \beta] -\log \pi)\right\}  \\
& =\mathbb{E}_{q(\beta)}[\beta] \sum_{n=1}^N \sum_{k=1}^K r_{nk}   \| \mX^n - g(\vh_n|\tW_{\it{dec}}^k)  \|_F^2 \\
&\,\,\,\,\,\,- \sum_{n=1}^N \sum_{k=1}^K r_{nk}\frac{T_nD}{2}(\mathbb{E}_{q(\beta)}[\log \beta] -\log \pi)  \\
&\propto \sum_{n=1}^N\sum_{k=1}^K r_{nk}   \| \mX^n - g(\vh_n|\tW_{\it{dec}}^k)  \|_F^2 + const.
\end{flalign*}
where we used $r_{nk} = \mathbb{E}_{q(\vy_n)}[y_{nk}]$.

We achieve this by applying RNN algorithm.
From the above discussion including Appendix \ref{sec:VB_algo}, we obtain the MDRA algorithm.

\subsection{Derivation of $\mathbb{E}_{Gamma(\beta|\nu,\lambda)}[\log \beta]$}
\label{DeriveEqlog}
By putting $\beta = {e}^x$, we obtain $x=\log \beta,d\beta = {e}^x dx$,
\begin{equation*}
\begin{split}
\mathbb{E}_{Gamma(\beta|\nu,\lambda)}[\log \beta] &= \int_0^\infty \log \beta \frac{\lambda^\nu}{\Gamma(\nu)}\beta^{\nu-1}e^{-\lambda x}d\beta \\
&= \int x \frac{\lambda^\nu}{\Gamma(\nu)}({e}^x)^{\nu-1}{e}^{-\lambda {e}^x} {e}^{x} dx \\ 
&= \int x \frac{\lambda^\nu}{\Gamma(\nu)}{e}^{x(\nu-1)}{e}^{-\lambda {e}^x} {e}^{x} dx \\
&= \int x \frac{\lambda^\nu}{\Gamma(\nu)}{e}^{x\nu-\lambda {e}^x} dx . \\
\end{split}
\end{equation*}
We here use 
\[
\frac{d}{d\nu} e^{x\nu-\lambda e^x} = xe^{x\nu-\lambda e^x},
\]
then the above equation is 
\begin{equation*}
\begin{split}
\mathbb{E}_{Gamma(\beta|\nu,\lambda)}[\log \beta] &=\int \frac{\lambda^\nu}{\Gamma(\nu)}\frac{d}{d\nu}{e}^{x\nu-\lambda {e}^x} dx \\
&= \frac{\lambda^\nu}{\Gamma(\nu)}\frac{d}{d\nu} \int {e}^{x\nu-\lambda {e}^x} dx .
\end{split}
\end{equation*}
In addition, $\int _0^\infty x^{\nu-1} {e}^{-\lambda {e}^x}dx$ is the normalization constant of gamma distribution, therefore
it equals to $\Gamma(\nu)/\lambda^\nu$. Hence we finally obtain
\begin{equation*}
\begin{split}
\mathbb{E}_{Gamma(\beta|\nu,\lambda)}[\log \beta] &= \frac{\lambda^\nu}{\Gamma(\nu)}\frac{d}{d\nu} \frac{\Gamma(\nu)}{\lambda^\nu}\\
&= \frac{\lambda^\nu}{\Gamma(\nu)} \frac{\Gamma ' (\nu) \lambda^\nu- \Gamma(\nu)\lambda^\nu \log \lambda }{\lambda^{2\nu}}\\
%&= \frac{\Gamma'(\nu)}{\Gamma(\nu)}-\log \lambda = \frac{d}{d\nu} \log \Gamma(\nu) -\log \lambda\\
& =\psi(\nu)-\log \lambda .
\end{split}
\end{equation*} 

\subsection{Parameter Setting}
\label{ParamSet}
In this section, we show the parameter setting of the experiments in Section \ref{Exp}.
% \usepackage{multirow}
\begin{table}[h]
\centering
\caption{Parameter setting (periodic signals)}
%\label{tab:my-table}
\begin{tabular}{|l|c|c|c|c|c|c|c|c|}
\hline
\multirow{2}{*}{}            & \multirow{2}{*}{$L$} & \multicolumn{3}{c|}{EUNN}                                    & \multicolumn{4}{c|}{VB}            \\ \cline{3-9} 
                             &                      & cap.               & fft                & cpx                & $K$ & $\theta_0$ & $\nu_0$ & $\lambda_0$ \\ \hline
\multicolumn{1}{|c|}{RNN-AE} & 4                   & \multirow{2}{*}{8} & \multirow{2}{*}{T} & \multirow{2}{*}{F} & -   & -        & -     & -         \\ \cline{1-2} \cline{6-9} 
\multicolumn{1}{|c|}{MDRA}   & 4                  &                    &                    &                    & 5   & 0.5     & 1.0   & 0.01       \\ \hline
\multicolumn{1}{|c|}{LSTM-AE}   & 8               & -              & -                   & -                  & -   & -        & -     & -       \\ \hline
\end{tabular}
\label{tbl:Priodic}
\end{table}

\begin{table}[h]
\centering
\caption{Parameter setting (complex periodic signals)}
%\label{tab:my-table}
\begin{tabular}{|l|c|c|c|c|c|c|c|c|}
\hline
\multirow{2}{*}{}            & \multirow{2}{*}{$L$} & \multicolumn{3}{c|}{EUNN}                                    & \multicolumn{4}{c|}{VB}            \\ \cline{3-9} 
                             &                      & cap.               & fft                & cpx                & $K$ & $\theta_0$ & $\nu_0$ & $\lambda_0$ \\ \hline
\multicolumn{1}{|c|}{MDRA}   & 4                  &   8                 & T                & F                   & 5   & 1.0     & 1.0   & 0.01       \\ \hline
\end{tabular}
\label{tbl:cpxPeridic}
\end{table}

\begin{table}[h]
\centering
\caption{Parameter setting (route clustering)}
%\label{tab:my-table}
\begin{tabular}{|l|c|c|c|c|c|c|c|c|}
\hline
\multirow{2}{*}{}            & \multirow{2}{*}{$L$} & \multicolumn{3}{c|}{EUNN}                                    & \multicolumn{4}{c|}{VB}            \\ \cline{3-9} 
                             &                      & cap.               & fft                & cpx                & $K$ & $\theta_0$ & $\nu_0$ & $\lambda_0$ \\ \hline
\multicolumn{1}{|c|}{MDRA}   & 4                  &   8                 & T                & F                    & 10   & 10.0     & 1.0   & 5.0       \\ \hline
\end{tabular}
\label{tbl:Route}
\end{table}

Here $L$ is the dimension of hidden variable $\bm{h}$, capacity, fft and cpx are parameters of EUNN \cite{Jing1}, $K$ is the number of the decoders,
$\theta_0,\nu_0,\lambda_0$ are hyperparameters of prior distributions.

\end{document}